\documentclass[12pt]{article}
\usepackage{amsmath}
\usepackage{times}
\usepackage{graphicx}
\usepackage{color}
\usepackage{multirow}
\usepackage[authoryear]{natbib}
\usepackage{rotating}
\usepackage{bbm}
\usepackage{latexsym}

\usepackage{amsmath}
\usepackage{amssymb}

\usepackage[printonlyused]{acronym}
\usepackage{mathtools}
\usepackage[bottom]{footmisc}

\newcommand{\captionfonts}{\normalsize}

\makeatletter  
\long\def\@makecaption#1#2{%
  \vskip\abovecaptionskip
  \sbox\@tempboxa{{\captionfonts #1: #2}}%
  \ifdim \wd\@tempboxa >\hsize
    {\captionfonts #1: #2\par}
  \else
    \hbox to\hsize{\hfil\box\@tempboxa\hfil}%
  \fi
  \vskip\belowcaptionskip}
\makeatother   

\begin{document}

\newacro{DL}[DL]{Deep-Learning}
\newacro{DNN}[DNN]{Deep Neural Network}
\newacro{ANNs}[ANNs]{Artificial Neural Networks}
\newacro{NNs}[\textit{NNs}]{\textit{Neural Networks}}
\newacro{NN}[\textit{NN}]{\textit{Neural Network}}
\newacro{TSFO}[TSFO]{Time-Stationary Frequency-Optimized}
\newacro{BP}[BP]{backpropagation}
\newacro{FP}[FP]{Forward-Propagation}
\newacro{CF}[CF]{Cost-Function}
\newacro{clCF}[$C_{c}$]{closed loop Cost-Function}
\newacro{clE}[$E_{c}$]{closed loop Error}
\newacro{ewG}[$G^{e}_{\omega}$]{closed loop Error Gradient}
\newacro{tCF}[$C_{t}$]{Traditional Cost-Function}
\newacro{tE}[$E_{t}$]{Traditional Error}
\newacro{oCF}[$C_{o}$]{open-loop Cost-Function}
\newacro{oE}[$E_{o}$]{open-loop Error}
\newacro{RDL}[RDL]{Reinforced Deep Learning}
\newacro{IDL}[IDL]{Inter-Domain Learning}
\newacro{rms}[RMS]{Root Mean Square}
\newacro{IDG}[$G^{c}_{\omega}$]{Inter-Domain Gradient}
\newacro{clEG}[$G_{A}^{E}$]{closed loop Error Gradient}
\newacro{clG}[$G^{c_{c}}_{a}$]{closed loop Gradient}
\newacro{nG}[$G^{a}_{\omega}$]{Network Gradient}
\newacro{tG}[$G^{c_{o}}_{a}$]{Traditional Gradient}
\newacro{oG}[$G^{c_{o}}_{a}$]{open-loop Gradient}
\newacro{iG}[$G$]{Internal Gradient}
\newacro{iE}[$\Phi$]{Internal Error}
\newacro{PG}[PG]{Partial Gradient}
\newacro{RL}[RL]{reinforcement learning}
\newacro{ARL}[ARL]{Analogue Reinforced Learning}
\newacro{LU}[$N{\omega}$]{Learning Unit}
\newacro{PID}[PID]{Proportional Integral Derivative}

\hspace{13.9cm}1

\ \vspace{20mm}\\

{\center{\Large{\sc{closed-loop deep learning: generating forward models with back-propagation\\}}}}
\ \\
{\bf \large Sama Daryanavard$^{\displaystyle 1},$  Bernd Porr$^{\displaystyle 1}$}\\
{$^{\displaystyle 1}$Biomedical Engineering Division, School of Engineering, University of Glasgow, Glasgow G12 8QQ, UK.}\\
%


\thispagestyle{empty}
\markboth{}{NC instructions}
\ \vspace{-0mm}\\
%
\begin{center} {\bf Abstract}
\end{center}
A reflex is a simple closed loop control approach which tries to
minimise an error but fails to do so because it will always react too
late. An adaptive algorithm can use this error to learn a forward
model with the help of predictive cues. For example a driver learns to
improve their steering by looking ahead to avoid steering in the last minute. In order to
process complex cues such as the road ahead deep learning is a
natural choice. However, this is usually only achieved indirectly by
employing deep reinforcement learning having a discrete state space. Here, we show how
this can be directly achieved by embedding deep learning into a closed
loop system and preserving its continuous processing. We show
specifically how error back-propagation can be achieved in z-space and
in general how gradient based approaches can be analysed in such closed
loop scenarios. The performance of this learning paradigm is
demonstrated using a line-follower both in simulation and
on a real robot that show very fast and continuous learning.

\section{Introduction}
Reinforcement learning \citep{Sutton98} has enjoyed a revival in recent years, reaching super human levels at video game playing
\citep{Guo2014}. Its success is owed to a combination of variants
of Q learning \citep{Dayan1992} and deep learning
\citep{Rumelhart1986}. This approach is powerful because deep learning
is able to map large input spaces, such as camera images or pixels of
a video game onto a representation of future rewards or threats
which can then inform an actor to create actions as to maximise such
future rewards. However, its speed of learning is still slow and
its discrete state space limits its applicability to robotics.

Classical control on the other hand operates in continuous time
\citep{Phillips2000} which potentially offers solutions to the
problems encountered in discrete action space. Adaptive approaches in
control develop forward models where an adaptive
controller learns to minimise an error arising from a fixed feedback
controller (for example \ac{PID} controllers) often called ``reflex''. This has been
shown to work for simple networks
\citep{Klopf86,Verschure91} where the error signal from the
feedback loop successfully learns forward models of simple predictive
(reflex-) actions. For example such a network was able to improve the
steering actions of a car were a non-optimal hard wired steering is
then quickly superseded by a forward model based on camera information
of the road ahead. Such learning is close to one shot learning in this
scenario because at every time step the error signal from the \ac{PID}
controller is available and adjusts the network
\citep{Porr2006ICO}. However, so far these networks could not easily be
scaled up to deeper structures and consequently had limited
performance \citep{Kulvicius2007}.

A natural step is to employ deep learning \citep{Rumelhart1986}
instead of a shallow network to learn a forward model. If we directly
learn a forward model with the deep network mapping sensor inputs to
actions then we no longer need a discrete action space. This then will
allow to potentially much higher learning rates because the error
feedback will be continuous as well. In order to achieve this we need
to define a new cost function for our deep network which is defined
within the closed loop framework benchmarking the forward model in
contrast to a desired output.

In this paper we present a new approach for direct use of deep
learning in a closed loop context where it learns to replace a fixed
feedback controller with a forward model. We show in an analytical way
how to use the Laplace/z-space to solve back-propagation in a closed
loop system. We then apply the solution first to a simulated line
follower and then to a real robot where a deep network learns fast to
replace a simple fixed \ac{PID} controller with a forward model.

\section{The learning platform}
\begin{figure}[htb]
\centering
\includegraphics[width=1\linewidth]{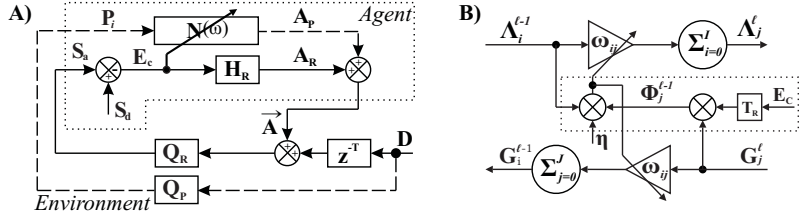}
\caption{\textit{\textbf{A)} The closed loop platform: consists of an inner reflex loop (solid lines) and an outer learning loop (dashed lines), the learning unit $N(\omega)$ generates a forward model of the environment. Given predictive inputs $U_{i}$ (filtered predictive signals $P_{i}F_{i}$) it generates an action $A_{P}$ that combats the disturbance $D$ on its arrival at the reflex loop. Finally, the closed loop error \acs{clE} gives an instructive feedback to the learning unit on how well $A_{P}$ protected the system from $D$. \textbf{B)} The computational unit: shows the forward and back-propagation of the inputs and the error to the deeper layers. Dotted line marks the correlation of the closed loop error with the internal parameters of the neuron highlighting the \textit{update rule}, where, $T_{R}$ is the transfer function of the reflex loop}}\label{fig:platform}
\end{figure}

Before we introduce and explore the deep learner $N$ we need to establish our
closed loop system. The configuration depicted in
Fig.~\ref{fig:platform}$A)$ is the architecture of this learning
paradigm which provides a closed loop platform for autonomous
learning. It consists of an inner reflex loop and an outer predictive
loop that contains the learning unit. In the absence of any learning,
the reflex loop receives a delayed disturbance $Dz^{-T}$ via the
reflex environment $Q_{R}$; leads to state $S_{a}$. Given
the desired state $S_{d}$ the \ac{clE} is generated as:
$\acs{clE}=S_{d}-S_{a}$. This drives the agent to take an appropriate
reflex action $A_{R}$ as to recover to $S_{d}$ and force
\ac{clE} to zero. However the reflex mechanism $H_{R}$ can
only react to the disturbance $D$ after it has perturbed the system.

Hence, the aim of the learning loop is to fend off $D$ before it
has disturbed the state of the robot. To that end, this loop receives
$D$ via the predictive environment $Q_{P}$ and in advance of
the reflex loop. This provides the learning unit with predictive
signals $P_{i}$ and, given its internal parameters $\omega$, a
predictive action is generated as: $A_{P}=N(P_{i},\omega)$.

During the learning process, $A_{P}$ combined with $A_{R}$ and
$Dz^{-T}$ travels through the reflex loop and \ac{clE} is
generated. This error signal provides the deep learner $N$ with a minimal
instructive feedback. Upon learning, $A_{P}$ fully combats $D$
on its arrival at the reflex loop (i.e. $Dz^{-T}$); hence the
reflex mechanism is no longer evoked and \ac{clE} is kept at zero.

\section{Closed loop dynamics}
The aim of the learning is to keep the \acl{clE} \acs{clE} to zero. Referring to Fig.~\ref{fig:platform}A this signal is derived as: $\ac{clE}(z)=S_{d}(z)-S_{a}(z)$; expansion of $S_{a}(z)$ yields:\footnote{For brevity, the complex frequency variable (z) will be omitted}
\begin{align}\label{eq:E}
\ac{clE}=S_{d}-Q_{R}(Dz^{-T}+\ac{clE}H_{R}+A_{P})=\frac{S_{d}-Q_{R}(Dz^{-T}+A_{P})}{1+H_{R}Q_{R}}
\end{align}
In mathematical terms, \textit{learning} entails the adjustment of the internal parameters of the learning unit $\omega$ so that \ac{clE} is kept at zero. To that end, the \acl{clCF} \acs{clCF} is defined as the square of absolute \ac{clE}: 
\begin{align}\label{eq:Cc}
\acs{clCF} \coloneqq |\acs{clE}|^{2}
\end{align}
Introduction of \ac{clCF} translates the \textit{learning goal} into adjustments of $\omega$ so that \ac{clCF} is minimised, this in turn ensures that \ac{clE} is kept at zero.
\begin{align}
\frac{\partial\ac{clCF}}{\partial \omega}=2 |\ac{clE}| \frac{\partial \ac{clE}}{\partial \omega} \Bigr\rvert_{\omega_{min}} =0 \left\{\begin{array}{l}\ac{clE}=0 \textit{, learning goal}\\
\ac{clE} \neq 0 \textit{, local minima} \end{array} \right.
\end{align}
The behaviour of the gradient $\frac{\partial\ac{clCF}}{\partial \omega} = \acs{IDG}$ is best explained through separation of gradients of the closed loop and the learner as below:
\begin{align}\label{eq:dCdw}
\acs{IDG} \coloneqq \frac{\partial\acs{clCF}}{\partial \omega} = \frac{\partial \acs{clCF}}{\partial A_{P}} \frac{\partial A_{P}}{\partial \omega} =\acs{clG} \acs{nG}
\end{align}
The former partial derivative, termed \acl{clG} \acs{clG}, solely relates to the dynamics of the closed loop platform; this is derived from equations~\ref{eq:E} and~\ref{eq:Cc}:
\begin{align}\label{eq:dcda}
\acs{clG} \coloneqq \frac{\partial \ac{clCF}}{\partial A_{P}} = 2|\ac{clE}| \frac{\partial \ac{clE}}{\partial A_{P}} = 2|\ac{clE}|\frac{-Q_{R}}{1+H_{R}Q_{R}} = 2|\ac{clE}| T_{R}
\end{align}
Where the resulting fraction $\frac{-Q_{R}}{1+H_{R}Q_{R}}$ is the transfer function of the reflex loop $T_{R}$.

\section{Towards closed loop error backpropagation}
To be able to link open loop backpropagation to our closed loop
learning paradigm we need to relate our closed loop error $\acs{clE}$ to
the standard open loop error of backpropagation. In conventional
open-loop implementations, the \acl{oCF} \acs{oCF} and \acl{oE}
\acs{oE} are defined at the action output of the network:
\begin{align}\label{eq:ct}
\acs{oCF} \coloneqq |\acs{oE}|^{2} \coloneqq |A^{d}_{P}-A_{P}|^{2}
\end{align}
Where $A_{P}^{d}$ is the desired predictive action. Minimisation
of \acs{oCF} with respect to the internal parameters of the
learning unit $\omega$ gives:
\begin{align}\label{eq:dcodw}
\frac{\partial \acs{oCF}}{\partial \omega} = \frac{\partial \acs{oCF}}{\partial A_{P}} \frac{\partial A_{P}}{\partial \omega} = \acs{tG}\acs{nG}
\end{align}
The former partial derivative is termed \acl{oG} \acs{oG}, from equation~\ref{eq:ct}:
\begin{align}
\acs{oG} = 2 \acs{oE} \frac{\partial \acs{oE}}{\partial A_{p}} =-2|A^{d}_{P}-A_{P}|
\end{align}
Now we relate the open-loop parameters to their closed loop counterparts.
Expansion of $S_{d}$ in equation~\ref{eq:E} gives:
\begin{align}
\ac{clE} &= Q_{R}(Dz^{-T}+\ac{clE}H_{R}+A_{P}^{d}) \text{ - } Q_{R}(Dz^{-T}+\ac{clE}H_{R}+A_{P}) = Q_{R}(A_{P}^{d}\text{ - }A_{P}) = Q_{R} \acs{oE}
\end{align}
Given $Q_{R}$ is a non-zero transfer function, the \acl{oE} is
kept at zero if and only if \acl{clE} is kept at zero:
\begin{align}
Q_{R} \neq 0 \quad \textit{ , therefore: } \quad \acs{clE} = 0 \iff \acs{oE}=0
\end{align}
Having now established how the error can be fed into an error backpropagation framework we are now
able to present the inner workings of the learning unit.

\section{The inner workings of the learning unit}
Having explored the dynamics of the closed loop, we now focus on the
inner working of the learning unit. The latter partial derivative in
equations~\ref{eq:dCdw} and~\ref{eq:dcodw}, termed the \acl{nG}
\acs{nG}, is merely based on the inner configuration of the learning
unit which in this work, is a \ac{DNN} with \ac{BP}. Given that the
network is situated in the closed loop platform, its dynamics is
expressed in z-space. The \ac{FP} entails feeding the predictive
inputs $P_{i}$ and generating the predictive action
$A_{P}$. This is shown in Fig.~\ref{fig:platform}$B)$ with solid
line and is expressed as below where $\Lambda_{j}^{\ell}$ denotes the activation of neurons\footnote{Subscripts refer to the neuron's index and superscripts refer to the layer containing the neuron or weight}:
\begin{equation}\label{frwdProp}
\Lambda_{j}^{\ell} = \Sigma_{i=0}^{I}\omega^{\ell}_{ij} \Lambda^{\ell-1}_{i} \quad \textit{ where: } \quad \ell:0 \rightarrow L \quad \textit{ note that: } \quad \Lambda^{-1}_{i}=P_{i}
\end{equation}
$\omega^{\ell}_{ij}$ denotes the weights of neurons which are analogous to weights in time-domain.
Using equation~\ref{frwdProp}, \acs{nG} with respect to specific weights gives:
\begin{align}\label{eq:partialCw}
G^{a}_{\omega_{ij}^{\ell}} \coloneqq\frac{\partial A_{P}}{\partial \omega_{ij}^{\ell}} = \frac{\partial A_{P}}{\partial \Lambda_{j}^{\ell}} \frac{\partial \Lambda_{j}^{\ell}}{\partial \omega_{ij}^{\ell}}= \frac{\partial A_{P}}{\partial \Lambda_{j}^{\ell }} \Lambda_{i}^{\ell-1}
\end{align}
The resulting partial derivative is termed \acl{iG} \acs{iG} and is calculated using \acl{BP}:
\begin{align}\label{eq:iG}
\acs{iG}_{j}^{\ell} \coloneqq \frac{\partial A_{P}}{\partial \Lambda_{j}^{\ell }} = \Sigma_{k=0}^{K}(w^{\ell +1}_{jk} \acs{iG}_{k}^{\ell+1}) \quad \textit{ where: } \quad \ell: L-1 \rightarrow 0 \quad \textit{ note that: } \quad \acs{iG}_{0}^{L}=1
\end{align}
Therefore, the \acl{iE} \acs{iE} of the neuron, measuring sensitivity of the \acl{clCF} with respect to its activation, is given as below; refer to equations~\ref{eq:dcda} and~\ref{eq:iG}:
\begin{align}\label{eq:iE}
\acs{iE}_{j}^{\ell} \coloneqq \frac{\partial \acs{clCF}}{\partial \Lambda_{j}^{\ell }} = \frac{\partial \acs{clCF}}{\partial A_{P}} \frac{\partial A_{P}}{\partial \Lambda_{j}^{\ell }} = 2 \acs{iG}_{j}^{\ell} |\acs{clE}|\frac{-Q_{R}}{1+H_{R}Q_{R}} 
\end{align}
The update rule can be expressed as the correlation of the internal error of the neuron with the activation of the previous neuron:
\begin{equation}
\Delta\omega^{\ell}_{ij}=\eta \acs{iE}^{\ell}_{j}(z)\Lambda^{\ell-1}_{i}(-z) \quad , \quad \eta \ll 1 \label{idlweightc}
\end{equation}
The small learning rate $\eta$ ensures that the time-dependant weight change is small. Note that Eq.~\ref{idlweightc}
results in a weight change in the time domain which is calculated in z-space and for that reason we call
this learning scheme Inter Domain Learning (IDL).

The gradient of the \ac{clCF} with respect to an arbitrary weight is
given as following, referring to equations~\ref{eq:dcda},
~\ref{frwdProp}, and \ref{eq:iE}:
\begin{align}
\frac{\partial \acs{clCF}}{\partial \omega ^{\ell}_{ij}} &=  \frac{\partial \acs{clCF}}{\partial \Lambda_{j}^{\ell}} \frac{\partial \Lambda_{j}^{\ell}}{\partial \omega_{ij}^{\ell}}
= \frac{-2 |\acs{clE}| Q_{R}}{1+H_{R}Q_{R}} \Sigma_{k=0}^{K}(w^{\ell +1}_{jk} \acs{iG}_{k}^{\ell+1}) \Lambda_{i}^{\ell-1}= \Sigma_{k=0}^{K}(w^{\ell +1}_{jk} \acs{iE}_{k}^{\ell+1}) \Lambda_{i}^{\ell-1}
\end{align}
This shows that the changes in \ac{clCF} with respect to an arbitrary weight
depends on the weighted internal error introduced in the adjacent
deeper layer. This is the propagation of \ac{clCF} into the deeper
layers and shows the backpropagation in the z-domain.

\section{Results}
The performance of our \ac{IDL} paradigm is tested using a
line-follower both in simulation and through experiments with a
real robot. The learning paradigm was developed into a bespoke
low-level C++ external library \citep{Daryanavard2019}.
The transfer function of the reflex loop $T_{R}$, derived in
Equation~\ref{eq:dcda}, is set to unity for the following results.
\subsection{Simulations} 
\begin{figure}[htb]
\centering
\includegraphics[width=1\linewidth]{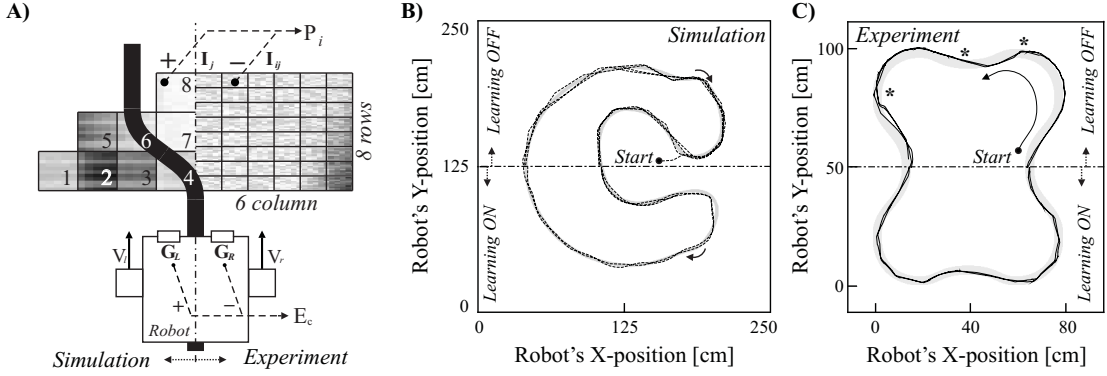}
\caption{\textit{\text{\bf{A)}} Schematic of the robot and its environment: the robot is composed of a body with two wheels with speeds of $V_{r}$ and $V_{l}$ and two ground sensors $G_{r}$ and $G_{l}$ from which the closed loop error \ac{clE} is generated. The robot is placed on a track and has vision of the path ahead. In simulations this is 16 symmetrical ground sensors (left-hand side) and in real experiments it is the camera view with 6x16 pixels (right-hand side). From this the predictors $P_{i}$ are generated as the difference of symmetrical pixels pairs. \text{\bf{B)}} and \text{\bf{C)}} The Track and robot's trajectory for simulations and experiments respectively. In both cases the top section shows the trajectory of the robot during a reflex trial showing a poor uneven trace whereas the bottom section shows the trajectory for a learning trial showing a smooth and even trace.}}\label{fig:tracking}
\end{figure}

\begin{figure}[htb]
\centering
\includegraphics[width=1\linewidth]{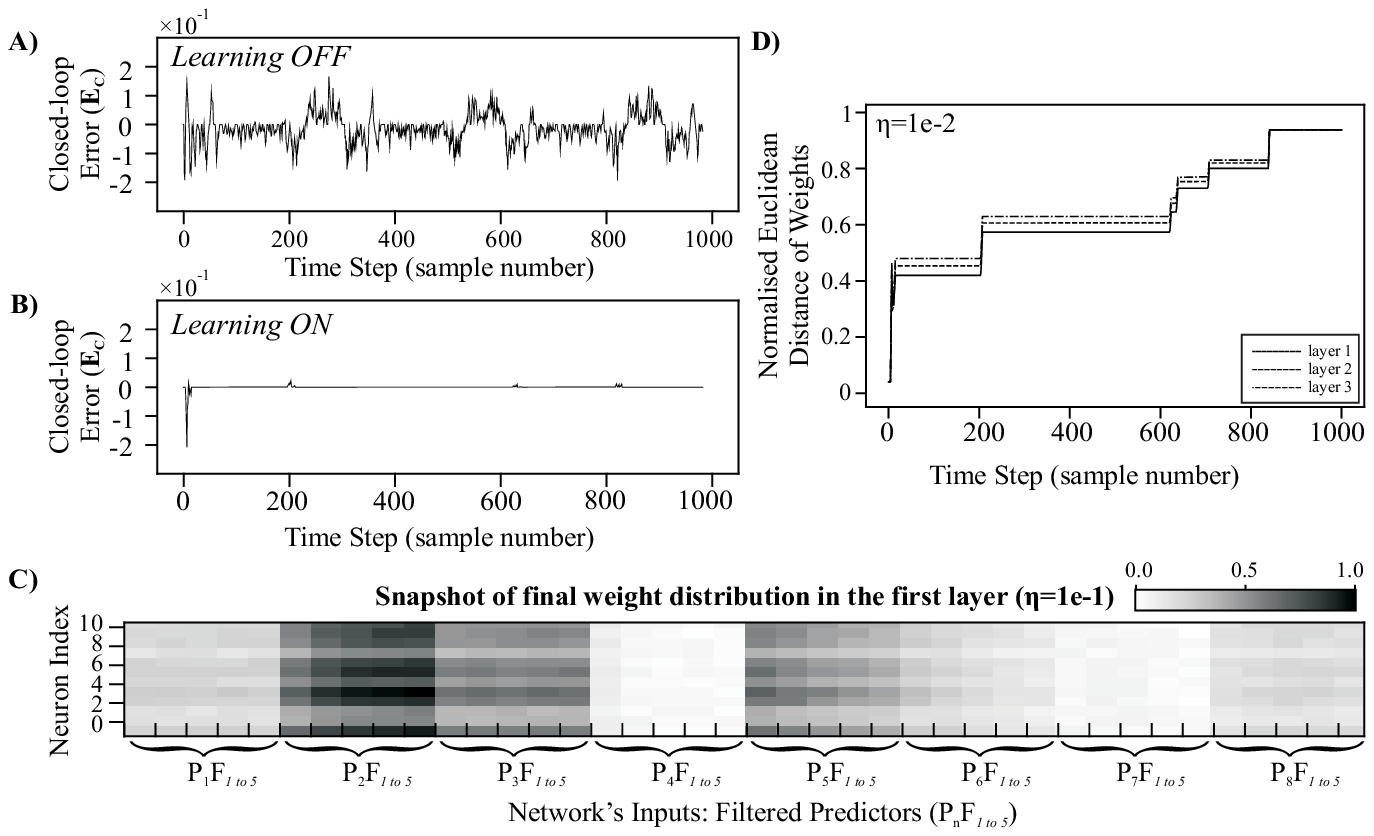}
\caption{\textit{Simulation results: \text{\bf{A)}} Shows robot's closed-loop error signal when navigating by reflex mechanism only, note the high amplitude (\acs{rms}=0.05) and frequent occurrence of the error whilst the learning is off. \text{\bf{B)}} Shows robot's closed-loop error signal when navigating by reflex mechanism and learning ($\eta = 10^{-2}$) mechanism, note that both the amplitude (\acs{rms}=0.01) and the occurrence of the error has reduced significantly with learning compared to that of reflex only. \text{\bf{C)}} Shows greyscale map of the weight distribution in the first layer after the learning is completed ($\eta = 10^{-2}$). Note that the weight distribution closely follow the location of predictors to which they associate, with weights associated to outermost predictors having high values and weights associated to innermost predictors having small values to allow for abrupt and subtle steering of the robot respectively. For position of predictors refer to Fig.~\ref{fig:tracking}A. \textbf{D)} Shows normalised euclidean distance of the weights in each layer during learning ($\eta = 10^{-2}$), note the gradual increase of the weight distance that stabilises towards the end of learning where the error is kept at zero.}}\label{fig:sim}
\end{figure}

Fig.~\ref{fig:tracking}A shows the configuration of the robot and its environment for simulations. The closed loop error \acs{clE} is calculated using the right and left ground sensors: 
\begin{align}
\ac{clE} = G_{L}-G_{R}
\end{align}
For prediction, 8 predictive signals are generated using an array of 16 ground sensors placed ahead of the robot as shown in the left-hand side of Fig.~\ref{fig:tracking}A. 
\begin{align}
P_{i}=I_{j}-I_{j^{*}} \quad \textit{, where $j^{*}$ is the sensor index symmetrical to j}
\end{align}
These are then filtered using a bank of 5 second-order lowpass filters ($f_{i}$), with a damping coefficient of $Q=0.51$ and impulse responses lasting between $3$ to $10$ iterations as to cause the correct delay for the correlation of predictors and the error signal. This results in 40 predictive inputs to the network which is configured with 12 and 6 neurons in the first two hidden layers and 1 output neuron in the final layer. The steering of the robot is facilitated through adjustments of the left and right wheel velocities (Fig.~\ref{fig:tracking}A):
\begin{align}
V_{R}=V_{0} + \alpha \ac{clE} + \beta \Lambda_{0}^{L} \quad \textit{, } \quad
V_{L} = V_{0} - \alpha \ac{clE} -\beta \Lambda_{0}^{L} \quad \textit{ and:  } \quad \vec{V} = \vec{V_{l}}+\vec{V_{r}}
\end{align}
Where $V_{0}$, $\alpha$ and $\beta$ are experimental tuning parameters set to $40$ \textit{\footnotesize{$[\frac{m}{s}]$}}, $200$ and $100$ respectively. 
The simulation environment is shown in Fig.~\ref{fig:tracking}B where the robot follows the track from the start point and in a loop for 1000 iterations. A set of simulations were carried out to contrast the reflex and the predictive behaviours; each scenario was repeated 10 times for reproducibility and statistical analysis.
Fig.~\ref{fig:sim}A and B show the average closed loop error over 10 trials for reflex and learning ($\eta = 1e{-2}$) behaviours respectively. A comparison of these results show very fast learning of the robot where the error signal is forced to zero. Top and bottom sections of Fig.~\ref{fig:tracking}B show the trajectory of the robot over the course of one trial, for the reflex and learning respectively; in the presence of learning the steering is of anticipatory nature and exhibits a smooth trajectory. Whereas, in the absence of learning the steering is  reactive and hence the abrupt response. 
\\

Fig.~\ref{fig:sim}D shows the normalised euclidean distance of the weights in each layer from their random initialisation. This shows a gradual increase from zero to its maximum during the course of one simulation. Since the error signal is propagated as a weighted sum of the internal errors all layers show similar rate of change in their weight distance.\\
Moreover, Fig.~\ref{fig:sim}C shows the final distribution of first layer's weights in the form of a normalised greyscale map upon completion of the learning. The weights show an organised distribution, with higher weights associated to the outer predictors, $P_{2,5}$ and smaller weights associated to the inner predictors, $P_{4,7,8}$; see Fig.~\ref{fig:tracking}A for position of predictors. This facilitates a sharper steering for the outer predictors ensuring a smooth trajectory, as shown in the bottom section of Fig.~\ref{fig:tracking}B.

\begin{figure}[htb]
\centering
\includegraphics[width=1\linewidth]{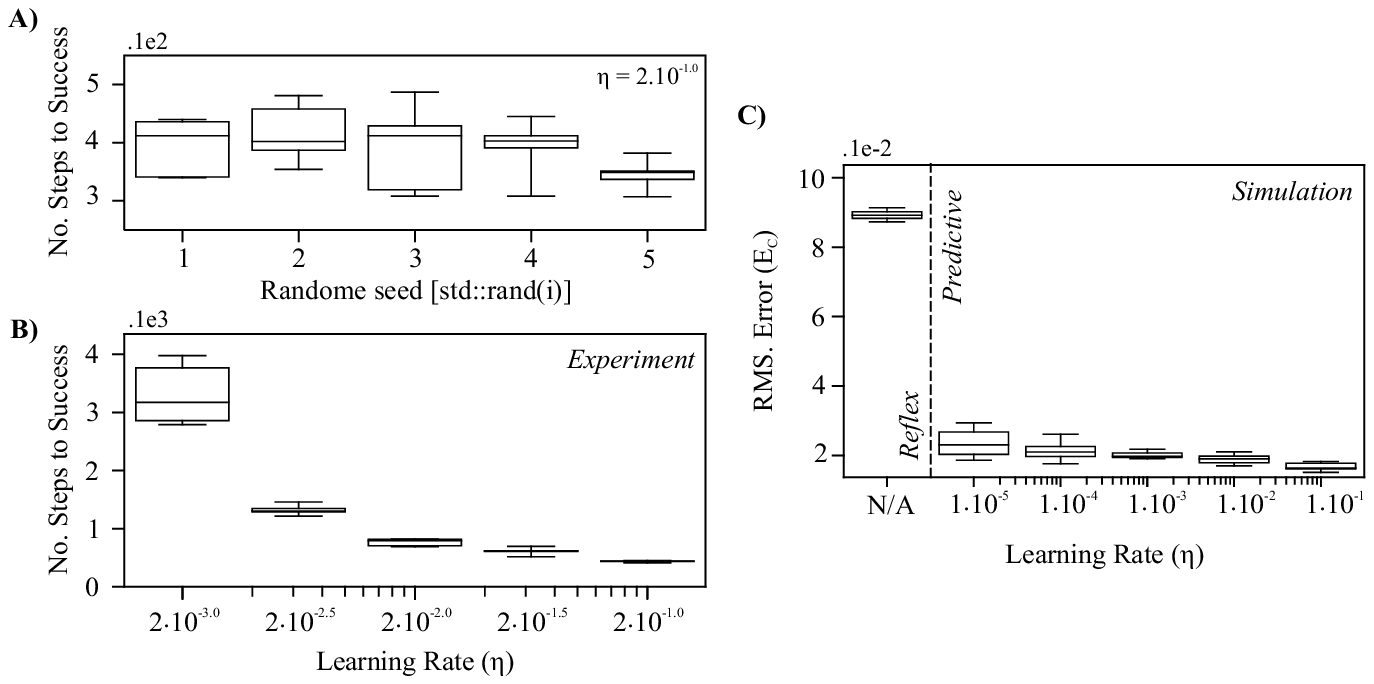}
\caption{\textit{\textbf{A)} Shows the number of steps
 taken until the success condition is met for 5 
 different random seed for weight initialisation.
   Note that the random initialisation of weights plays
    no significant role in the learning and success time.
     \textbf{B)} Shows the effect of learning rate on the
      number of step taken until the success condition is 
      met. Data shows a significant exponential decrease 
      in the time taken before a successful learning is 
      achieved. In other words, the learning is significantly
       faster for higher learning rates as it varies from
        $2.10^{-3}$ to $2.10^{-1}$. \textbf{C)} Shows the effect
       of learning rate on \ac{rms} value of closed-loop
        error \ac{clE}. Note the significant reduction of
         the closed-loop error in the presence of learning
          compared to that of reflex only, as well as the
           gradual improvement of learning (faster learning)
            with exponential increase of the learning rate 
            $\eta $ from $10^{-5}$ to $10^{-1}$. Examples of
             these trials are shown in Fig.~\ref{fig:sim}A and B 
for reflex and learning with $\eta=10^{-2}$ respectively.}}\label{fig:box}
\end{figure}

Another set of simulations were carried out with five orders of learning rates: $\eta:\{10^{-5},10^{-4},10^{-3},10^{-2},10^{-1}\}$; each of the scenarios were repeated 10 times. Fig.~\ref{fig:box}C shows the \ac{rms} of the error signal for each learning trials as well as that of the reflex trials for comparison. All learning scenarios show a significantly smaller \ac{rms} error when compared to the reflex behaviour; the error is reduced from over $8\cdot 10^{-2}$ to around $2\cdot 10^{-2}$ and below. There is a gradual decrease in this value as the learning rate is increased. Smaller values of \ac{rms} error indicates both the reduction in the amplitude and also the recurrence of the error signal.

\subsection{Experiments}
The experiments with a real robot were carried out using a Parallax
SumoBot as a mechanical test-bed, a Raspberry Pi 3B+ for computation and an Arduino Nano as the motor
controller. For predictive learning a camera was mounted on the robot providing vision of the path ahead (see right-hand side of Fig.~\ref{fig:tracking}) as a matrix of pixels,
$[I]_{NM}$, from which the predictive signals $P_{ij}$ are extracted :
\begin{align}
P_{ij}=I_{ij}-I_{ij^{*}} \quad \textit{, where $j^{*}$ is the sensor index symmetrical to j}
\end{align}
With 6 columns and 8 rows, 48 predictive signals were
extracted and filtered using 5 second-order low-pass filters with
$Q=0.51$ and impulse responses lasting from $5$ to $10$ time steps to
cause the correct delay. The error signal was
defined as a weighted sum of 3 light sensors for a smoother and more
informative error signal. The deep neural network was configured with 11 hidden layers each with 11 neurons, as well as an output layer with 3 neurons to facilitate slow, medium and sharp steering.

\begin{figure}[htb]
\centering
\includegraphics[width=1\linewidth]{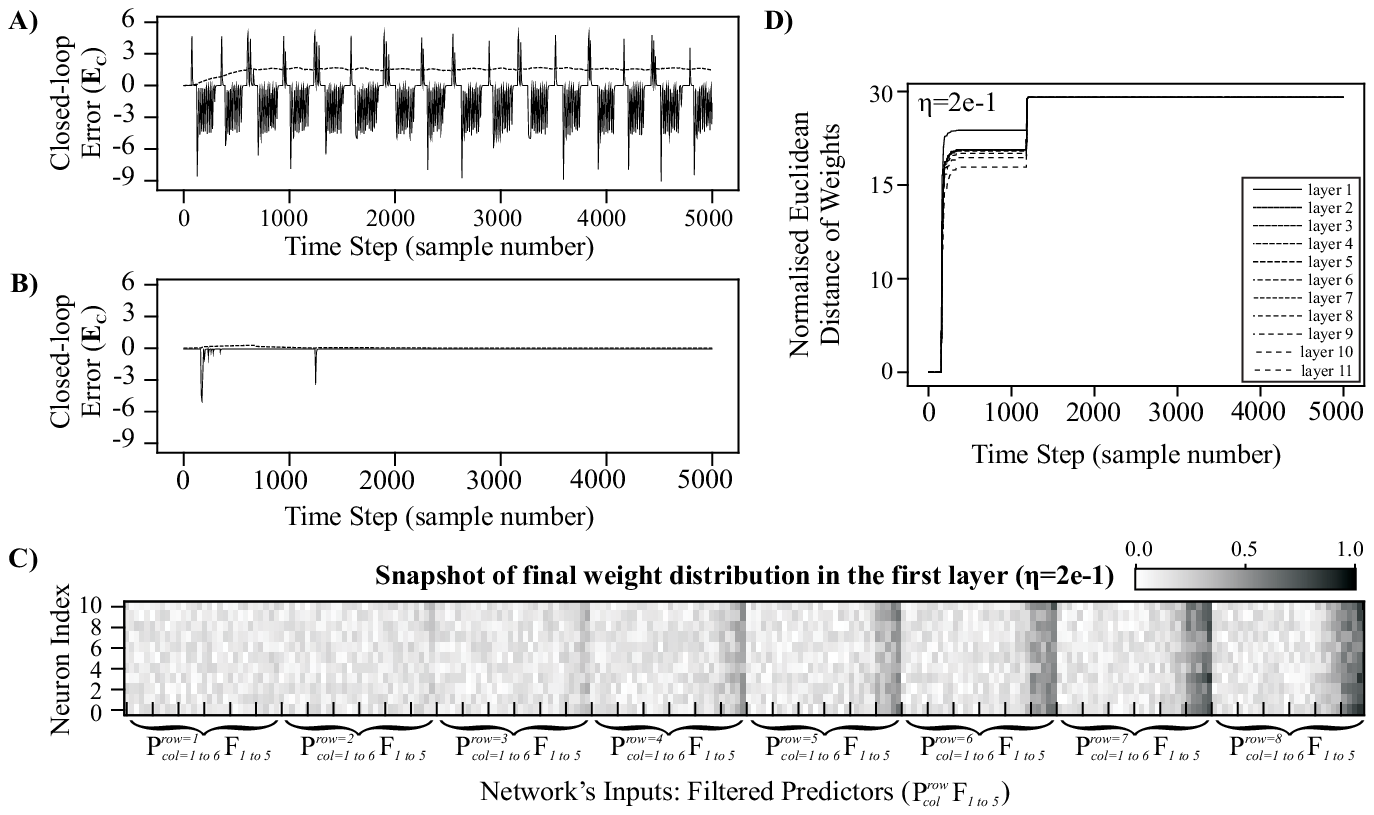}
\caption{\textit{Experimental results for learning rate of $\eta = 2.10^{-1}$: 
\textbf{A)} Shows the closed-loop error when robot navigates with 
reflex mechanism only. This sets a benchmark for evaluating the 
performance of the learning; note the high amplitude and persistence
 of this signal. \textbf{B)} Shows the closed-loop error when
  the learning mechanism governs the navigation of the robot.
   Note the significant reduction of the error signal compared 
   to the reflex data showing fast learning. 
   \textbf{C)} Shows a greyscale map of the 
  weight distribution in the first layer after the learning
   has been completed. Note that the weight distribution closely 
   follows the location of predictors to which they associate; 
   with weights associated to outermost predictors having 
   high values and weights associated to innermost 
   predictors having small values to allow for abrupt 
   and subtle steering of the robot respectively. 
   This greyscale mapping is also illustrated in 
   Fig.~\ref{fig:tracking}A. \textbf{D)} Shows 
   normalised euclidean distance of the weights 
   in each layer during learning.}}\label{fig:realh}
\end{figure}

Fig.~\ref{fig:realh}A Shows the closed-loop error in the absence of learning. This is when the robot navigates using its reflex system only. It can be seen that the error signal is very persistent in this case as the robot can only generate an appropriate steering command retrospectively after an error has occurred. This sets a benchmark for evaluation of the deep learner. Fig.~\ref{fig:realh}B shows the error signal in the presence of the deep learner where the
learning rate is $ \eta = 2\cdot 10^{-1}$, this shows a strong
reduction of the error signal over the first 1500 steps, where the learning is achieved rapidly using the closed-loop error that acts as a minimal instructive feedback for the deep learner. Fig.~\ref{fig:realh}C shows the final distribution of the weights in the first layer associating different strength to different pixel location of the predictors. From the gradient it can be seen that the farther the predictor from the centre line the greater the steering action, this is also illustrated in Fig.~\ref{fig:tracking}A. Fig.~\ref{fig:realh}D shows the weight
change in each layer as explained in the simulation results. The
weight distance changes noticeably over the first 1500 steps dictated by the closed-loop error but comes to a stable plateau as the error signal remains at zero.

Fig.~\ref{fig:tracking}C shows the robot's track and compares the trajectory of the robot for a reflex and a learning trial. Top section of the figure shows that when the learning is off the trace of robot almost always remains outside of the track with a few crossover points indicated by a star. Whereas, the bottom section shows that with learning ($\eta = 2.10^{-1}$) the trace of robot is aligned with the track.

The performance of the deep learner ($\eta = 2\cdot 10^{-1}$) was repeated with 5 different random weight initialisation using different random seeds $ srand(i)$ where $i = \{0,1,2,3,4\}$. In the presence of learning, "success" refers to a condition where the closed-loop error shows a minimum of 75 percent reduction from its average value during reflex only trials,  for 100 consequent steps. Fig.~\ref{fig:box}A shows that different random initialisation of the weights makes no significant difference to the time that it takes for the learner to meet the success condition.

The experiment was repeated with a 5 different learning rates
$\eta:\{2.10^{-3},2.10^{-2.5},$ $2.10^{-2},2.10^{-1.5},2.10^{-1}\}$  ; each experiment was repeated 5 times for reproducibility. Fig.~\ref{fig:box}B shows the time taken to success for these trials. This data shows an exponential decay of the success time as the learning rate is increased.

\begin{figure}[htb]
\centering
\includegraphics[width=1\linewidth]{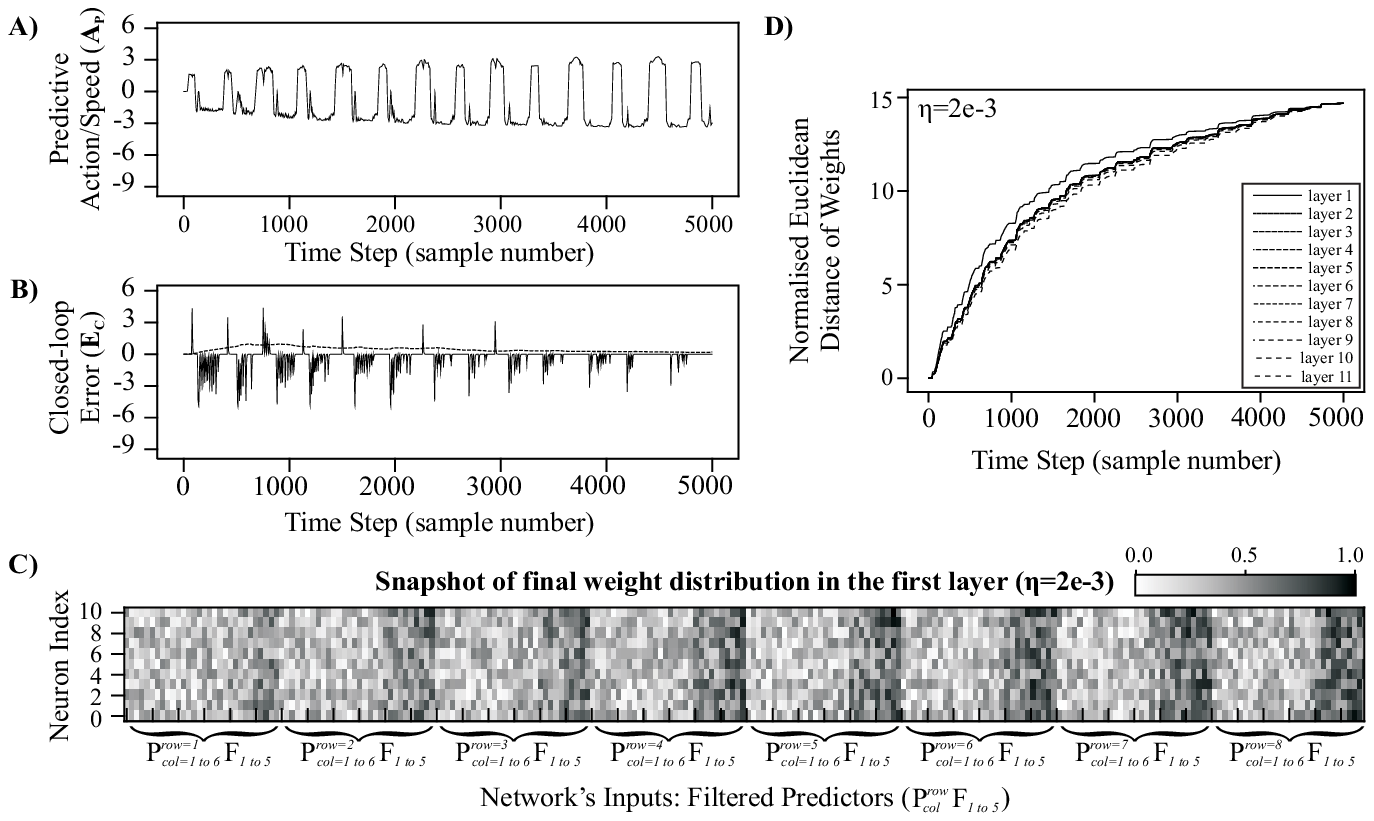}
\caption{\textit{Experimental results for learning rate of $\eta = 2.10^{-3}$: \textbf{A)} Shows the predictive action of the network $A_{P}$. This this the contribution of the learning to the steering of the robot is anticipation of a disturbance (turn in the road). Note that as the learning improves the amplitude of the steering increases and becomes more precise. \textbf{B)} Shows the closed-loop error when the learning mechanism governs the navigation of the robot. Note that the error is continuously reduced over time as the learning progresses. \textbf{C)} Shows a greyscale map of the weight distribution in the first layer after the learning is completed. Note that the weight distribution closely follow the location of predictors to which they associate, with weights associated to outermost predictors having high values and weights associated to innermost predictors having small values to allow for abrupt and subtle steering of the robot respectively. This greyscale mapping is also illustrated in Fig.~\ref{fig:tracking}A. \textbf{D)} Shows normalised euclidean distance of the weights in each layer during learning.}}\label{fig:reall}
\end{figure}

Fig.~\ref{fig:reall} shows another example of a learning trial similar to that in Fig.~\ref{fig:realh} but with a smaller learning rate $\eta = 2.10^{-3}$. Fig.~\ref{fig:reall}A shows the contribution of the deep learner to the resultant differential speed of the robot. This quantity is small and inaccurate at the start of the trial where the reflex mechanism governs the navigation of the robot, however, the contribution of the learner grows larger and more precise over time as the learner begins to dominate the navigation. This transition from reflex to learning navigation is also seen in Fig.~\ref{fig:reall}B where the error signal \ac{clE} decreases gradually as a successful learning is approached.
Fig.~\ref{fig:realh}C shows the final distribution of the weights in the first layer and shows similar but more crude gradients compared to that of Fig.~\ref{fig:reall}C. Fig.~\ref{fig:realh}D shows the weight
change in each layer during learning. This shows a more gradual change compared to the learning trial with $\eta = 2.10^{-1}$ shown in Fig.~\ref{fig:realh}D.

\section{Discussion}
In this paper we have presented a learning algorithm which creates a
forward model of a reflex employing a multi layered network. Previous
work in this area used shallow \citep{Kulvicius2007}, usually single layer networks to learn
a forward model \citep{nakanishi2004feedback,Porr2006ICO} and it was
not possible to employ deeper structures. On the other hand model free
RL has been using more complex network structures such as deep
learning by combining it with Q-learning where the network learns to
estimate an expected reward \citep{Guo2014,Bansal2016}. At first sight
this looks like two competing approaches because they both use deep
networks with error backpropagation. However, they serve different
purposes as discussed in \citet{dolan2013goals,Botvinick2014} which
lead to the idea of hierarchical RL where RL provides a prediction
error for an actor which can then develop forward models.

Both, in deep RL \citep{Guo2014} and in our algorithm we employ error
backpropagation which is a mathematical trick where an error/cost
function is expanded with the help of partial derivatives
\citep{Rumelhart1986}. This approach is appropriate for open loop
scenarios but for closed loop one needs to take into account the
endless recursion caused by the closed loop. In order to solve this
problem we have switched to the z-domain in which the
recursion turns into simple algebra. A different approach has been
taken by LSTM networks where the recursion is unrolled and
backpropagation in time is used to calculate the weights \citep{Hochreiter1997}
which is done offline whereas in our algorithm this is done while
the agent acts in its environment.

Deep learning is generally a slow learning algorithm and deep RL tends
to be even slower because of the sparsity of the discrete rewards. On
the other hand purely continuous or sampled continuous systems can be
very fast because they have continuous error feedback so that in terms
of behaviour nearly one shot learning can be achieved
\citep{Porr2006ICO}. However, this comes at the price namely that
forward models are learned from simple reflex behaviours and no
sophisticated planning can be achieved. For that reason it has been
suggested to combine the model free deep RL with model based learning
to have a slow and a fast system \citep{botvinick2019reinforcement}.

Forward models play an important role in robotic and biological motor control 
\citep{Wolpert1998,Wolpert2001,Haruno2001,nakanishi2004feedback} where
forward models guarantee an optimal trajectory after learning and
with our approach this offers opportunities to learn more complex
forward models with the help of deep networks and then combine it
with traditional Q-learning to planning those movements.

\bibliographystyle{plainnat}
\bibliography{myReff,embodiment,fcl,hebb,laplace,limbic,bernds,selforg}

\end{document}